\documentclass{article}

\usepackage{arxiv}

\usepackage[utf8]{inputenc} 
\usepackage[T1]{fontenc}    
\usepackage{hyperref}       
\usepackage{url}            
\usepackage{booktabs}       
\usepackage{amsfonts}       
\usepackage{nicefrac}       
\usepackage{microtype}      
\usepackage{lipsum}		
\usepackage{graphicx}
\usepackage{natbib}
\usepackage{doi}

\usepackage[frozencache,cachedir=.]{minted}

\usepackage{authblk}

\usepackage{caption}
\usepackage{subcaption}

\usepackage{tcolorbox}
\tcbuselibrary{minted,breakable,xparse,skins}

\definecolor{bg}{gray}{0.95}

\setminted[python]{breaklines, framesep=2mm, fontsize=\footnotesize, numbersep=5pt}

\usepackage{mdframed}

\usepackage{listings}
\usepackage{xcolor}
\usepackage{cancel}
\usepackage{soul}
 
\usepackage{comment}

\title{iquaflow: A new framework to measure image quality}

\author[1]{P. Gallés}
\author[1]{K. Tak\'ats}
\author[2]{M. Hernández-Cabronero}
\author[3]{D. Berga}
\author[1]{L. Pega}
\author[1]{L. Riordan-Chen}
\author[1]{C. Garcia}
\author[1]{G. Becker}
\author[3]{A. Garriga}
\author[3]{A. Bukva}
\author[2]{J. Serra-Sagristà}
\author[1]{D. Vilaseca}
\author[1]{J. Marín}

\affil[1]{Satellogic Inc; \texttt{pau.galles@satellogic.com}}
\affil[2]{Universitat Autònoma de Barcelona - UAB-DEIC-GICI; \texttt{miguel.hernandez@uab.cat}}
\affil[3]{EURECAT - Multimedia Technologies Unit; \texttt{david.berga@eurecat.org}}

\renewcommand{\shorttitle}{\textit{arXiv} Template}

\hypersetup{
pdftitle={iquaflow: A new framework to measure image quality},
pdfauthor={P. Gallés, K. Takáts, M. Hernández-Cabronero, D. Berga, L. Pega, G. Becker, J. Serra-Sagristà, D. Vilaseca, and J. Marín},
pdfkeywords={image quality, vision, deep learning, augmentation, compression},
}

\begin{document}

\maketitle

\begin{abstract}
\textsc{iquaflow} is a new image quality framework that provides a set of tools to assess image quality. The user can add custom metrics that can be easily integrated. Furthermore, \textsc{iquaflow} allows to measure quality by using the performance of AI models trained on the images as a proxy. This also helps to easily make studies of performance degradation of several modifications of the original dataset, for instance, with images reconstructed after different levels of lossy compression; satellite images would be a use case example, since they are commonly compressed before downloading to the ground. In this situation, the optimization problem consists in finding the smallest images that provide yet sufficient quality to meet the required performance of the deep learning algorithms. Thus, a study with \textsc{iquaflow} is suitable for such case. All this development is wrapped in \textsc{Mlflow}: an interactive tool used to visualize and summarize the results. This document describes different use cases and provides links to their respective repositories. To ease the creation of new studies, we include a cookiecutter\footnote{https://github.com/satellogic/iquaflow-use-case-cookiecutter} repository. The source code, issue tracker and aforementioned repositories are all hosted on GitHub \footnote{https://github.com/satellogic/iquaflow}.
\end{abstract}


\section{Introduction}

The increasing interest and investment in low-cost Earth Observation (EO) satellites (such as nanosats and microsats) in recent years has made possible the creation and improvement of multiple applications that feed on the images obtained \cite{buchen2014spaceworks}. Thanks to the emergence of new sensors, the quality of these images has increased significantly, thus contributing to the accuracy and efficiency of the applications that use them, giving rise to the NewSpace era, which has led to the emergence of many new companies. 

It is the case of Satellogic, a company that was founded in 2010 and specializes in Earth observation data and analytical imagery  solutions. Satellogic  designs,  builds  and  operates  its  own  fleet  of  Earth observation satellites to frequently collect affordable high-resolution imagery for decision-making in a broad range of industrial, environmental and government applications. The Satellogic satellite constellation consists of individual small satellites, named NewSats. Each of the NewSat satellites has a multispectral and a hyperspectral sensor. Its data is used in some of the studies mentioned in the present article.

Imagery is a means to an end. Large-scale data analytics and artificial intelligence equipment turns imagery into answers to help industries, governments and individuals solve problems, facilitate decision making and generate competitive advantage. In this context, the growth in the number of EO users has also increased the land area of interest and thus the amount of imagery required to meet their needs. As a result, an increasing volume of EO image data needs to be transmitted. However, the transmission capacity between satellites and ground stations has not grown at the same rate as the required volume of imagery.

Due to their orbit, satellites can only contact these stations for limited periods of time and with limited bandwidths. Moreover, the total amount of energy available for all satellite tasks - including image capture, processing and transmission - is also limited. These limitations pose a bottleneck in the operation of EO satellites since the quantity and quality of images reaching the ground is determined by the efficiency of transmission. In turn, this limitation has a negative impact both on the costs of EO services and on users and applications that increasingly demand higher quality and frequency of observation of the regions of interest. 

Given the aforementioned needs, this project seeks to optimize the decision making process when selecting an image processing algorithm to optimize the storage, quality and transmission of images either on EO satellites or on the ground. Optimization refers to finding and detecting the satellite image parameters that allow the smallest compressed data volume that provides sufficient quality to meet the required performance of the deep learning algorithms used. In order to achieve this objective a new framework named \textsc{iquaflow} (acronym of Image Quality Assessment) is proposed. The framework includes the necessary tools to draw conclusions based on specific metrics.

\begin{figure}[ht]
	\centering
	\includegraphics[width=10.5 cm]{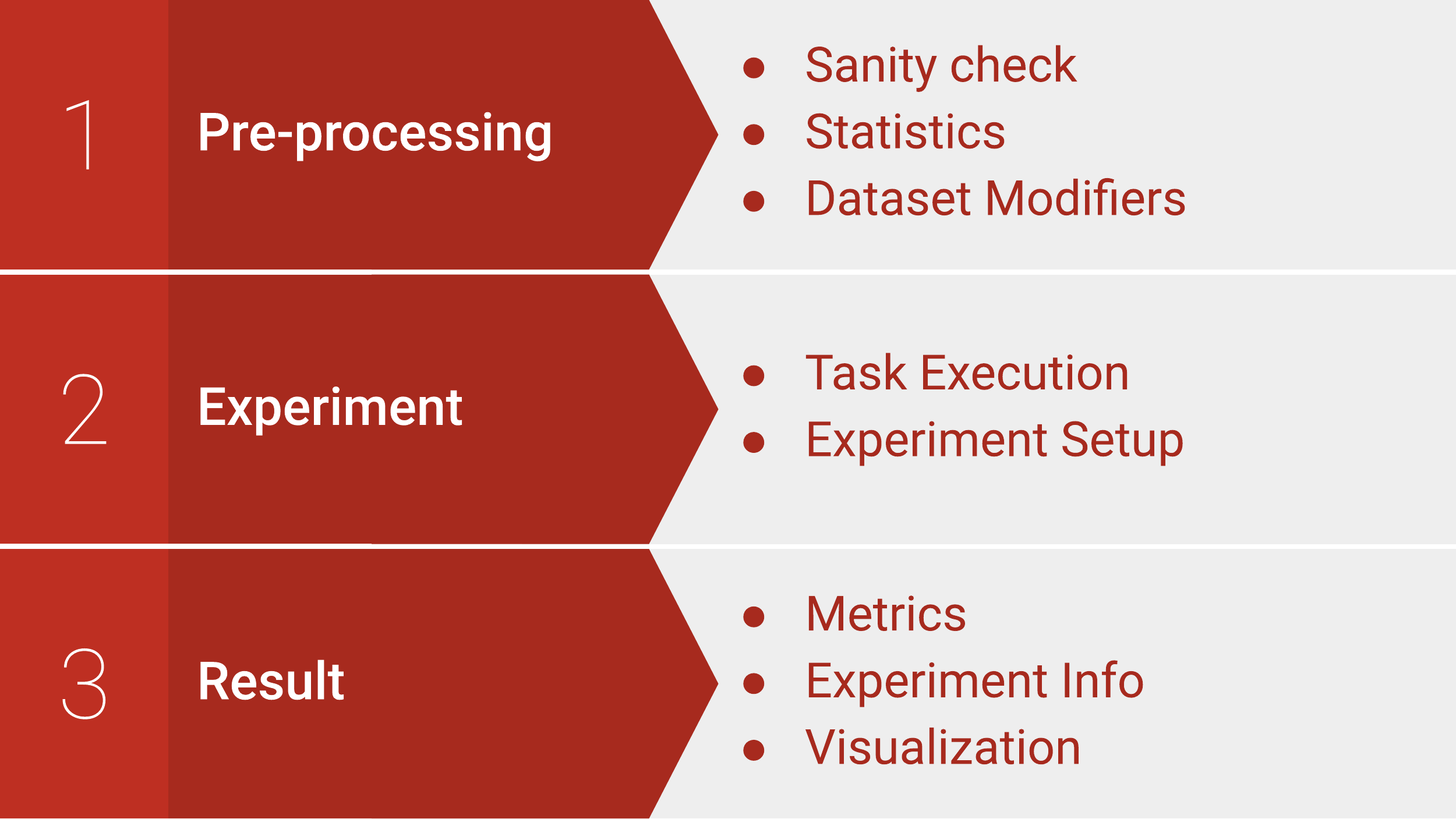}
	\caption{Typical use case workflow using \textsc{iquaflow}. The first step is to overview and sort the data. Then some modifiers can be defined. Afterwards, the task and the overall experiment is set and executed. Then additional metrics can be estimated on the results. Finally, one can summarize, sort and visualize the results.}
	\label{fig:workflow}
\end{figure}

\textsc{iquaflow} consists of multiple Python modules that can be imported in any image quality-related project for research or production. The framework also includes the necessary tools to automatize and organize experiments. Most of similar tools help to implement better tractability model trainings in order to perform model selection or model performance degradation studies (See \cite{Lofqvist2021}, \cite{Jo2021ImpactOI} and \cite{Delac2005}). Instead, \textsc{iquaflow}, analyzes the change in performance of the models based on modifications in quality of the training images. Having this perspective in mind, \textsc{iquaflow} organizes training executions based on qualities of the images. Thus, this approach is considered image quality selection rather than the common practice of image selection. Additionally, this framework provides a variety of tools that help to speed up the machine learning development such as dataset sanity check, dataset statistics, dataset visualization and dataset modification algorithms. It is designed to easily adapt to conventions. The tool can be imported into any kind of deep learning framework such as tensorflow, keras or pytorch. This allows to effortlessly generate experiments on any existing project regardless of its dependencies. Internally, this project relies in \href{https://mlflow.org/}{\textsc{Mlflow}}, which enables to organize the information locally or even in a remote server with an interaction that is abstracted to the user. As a result, the user does not need to learn another cumbersome framework. Ultimately, \textsc{iquaflow} only needs logging parameters that are easy to adapt (such as generating a json or saving files in a specific folder that is indicated as an input argument in the user script). Figure~\ref{fig:workflow} shows the typical workflow of a study made with \textsc{iquaflow}.

\section{Use cases}

\textsc{iquaflow} has already been used in various use cases to solve real problems that are faced in the EO industry. In this section we summarize some of its use cases highlighting the benefits of using \textsc{iquaflow} in them. The first section (\ref{mnistshowcse}) is actually an example of \textsc{iquaflow} usage rather than a real case. Then the other two sections (\ref{obbusecase} and \ref{superusecase}) are describing two groups of use cases related with object detection and super resolution respectively.

\subsection{MNIST showcase} \label{mnistshowcse}

The objective of this study \footnote{\href{https://github.com/satellogic/iquaflow-mnist-use-case}{https://github.com/satellogic/iquaflow-mnist-use-case}} is to examine how the training performance of a deep learning classifier degrades as the amount of noise in the input dataset is increased.  The dataset used is MNIST \cite{deng2012mnist}, which is widely used for benchmaking deep learning algorithms. It consists of a dataset of handwritten digits ($0$-$9$) images that is used to evaluate machine learning pattern recognition algorithms. Each digit image is $28~\times~28$ pixels.

The first step in the development process is to prepare the user training script that contains the deep learning classifier. In this case the model is built using the architecture of a resnet18 \cite{HeZRS15}. Afterwards the user script is adapted to follow the \textsc{iquaflow} conventions. This means to add the appropriate input arguments (See section \ref{Conventions}). The output that generates the user script also had to follow the standards of \textsc{iquaflow}. Next step is to create a custom modifier that integrates with \textsc{iquaflow} and does the desired alteration on the original dataset. In this case the modifier is adding noise following a gaussian distribution. The standard deviation of that noise distribution is an adjustable parameter. Finally \textsc{iquaflow} executed all requested combinations: hyperparameter variations, number of repetitions and dataset modifiers.

\begin{figure}[ht]
	\centering
	\includegraphics[width=10.5 cm]{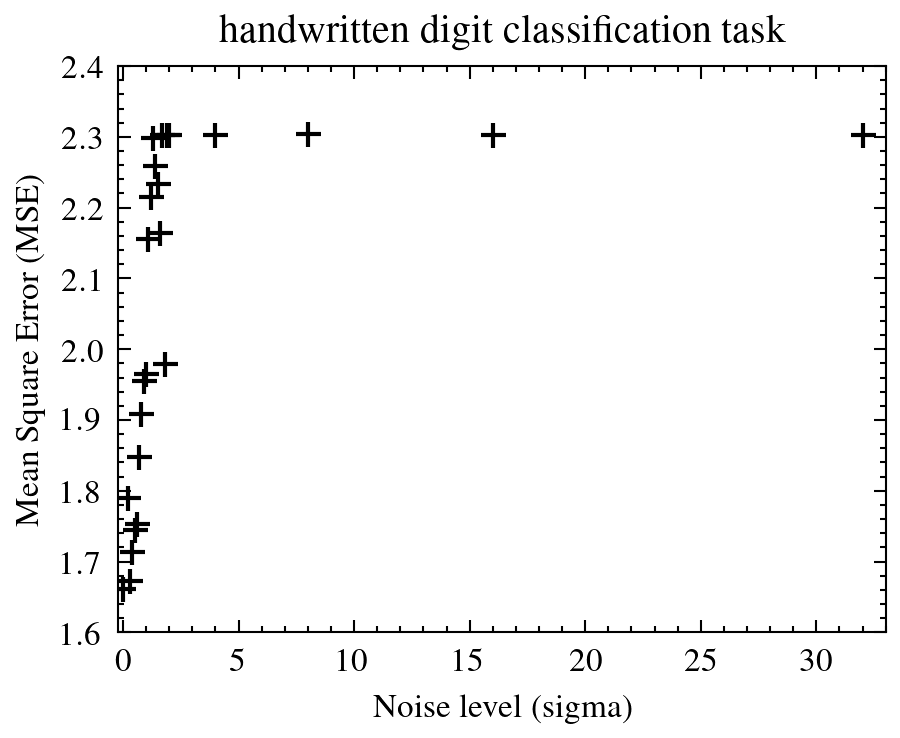}
	\caption{Mean Squared Error of the predicted digital label number with respect to the actual target number for different noise level dataset modifications. The noise level is defined as sigma of the Gaussian random noise distribution applied on them.}
	\label{fig:handwrittenclass}
\end{figure}

Despite being a classification task the loss function was set as Mean Squared Error with the aim to penalize further the predicted digits that are more distant from the actual labeled numbers. The Figure~\ref{fig:handwrittenclass} shows how the performance in the validation set degrades as the noise amount is increased up to a point where it does not degrade further (around sigma noise modifier of 2).

\subsection{Object Detection on compressed satellite images} \label{obbusecase}

As mentioned in the introduction EO satellites capturing images have limited energy resources. Further to that they have limited connection time and data flow capacity between the orbiting equipment and ground station systems on earth. The images are therefore compressed before downloading. There are compression algorithms that can be reverted recovering the original uncompressed image (lossless algorithms). On the other hand there are lossy algorithms that will compress further with irreversible operations (such as interpolating the image to a smaller size with less pixels ). When the usage of these images is well known the compression can be adjusted. This way, we avoid loss in performance on its intended application. This can be the case of detecting objects with a deep learning algorithm on satellite images.

We have built three use cases to study how the the application of different compression techniques affects object detection performance. In the first use case \footnote{\href{https://github.com/satellogic/iquaflow-dota-use-case}{https://github.com/satellogic/iquaflow-dota-use-case}}, we sought to replicate the study from \cite{Lofqvist2021} where they study the performance of CNN-based object detectors on constrained devices by applying different image compression techniques to satellite data. In this case they focus on execution times, memory consumption and some insights about accuracy. In our case we have focused on the performance metrics as a function of compression ratios. For this sudy the public dataset DOTA \cite{Xia_2018_CVPR} was used. The images were collected from different sensors with image sizes ranging from $800~\times~800$ to $20000~\times~20000$ pixels while the pixel size varies from $0.3$~m to $2$~m resolution. DOTA has different versions, in the present study DOTA-v1.0 has been used which contains 15 common categories, $2806$ images and more than $188$k object instances. The proportions of the training, validation, and testing sets in DOTA-v1.0 are $1/2$, $1/6$, and $1/3$ \cite{Xia_2018_CVPR}. A disadvantage of this dataset is that the test set is not openly available, rather it is in a form of a remote service to query the predictions. This does not allow to compress the test images the same way the other partitions are modified in the present study. To overcome this problem we divided up the validation set: half of it is used as actual validation and the other half for testing. Then the images are cropped to $1024~\times~1024$ with padding when necessary. After this operation the amount of crops for the partitions train, validation and testing are respectively $9734$, $2670$ and $2627$.

Our second use case\footnote{\href{https://github.com/satellogic/iquaflow-dota-obb-use-case}{https://github.com/satellogic/iquaflow-dota-obb-use-case}} is similar to the first one with newer implementations of object detection algorithms that are based on oriented annotations. The dataset used was the same as in the first study because the original annotations are oriented.

The final objective was to find an optimal compression ratio which is defined as the minimum average file size that can be set without lowering the performance. We have found this to be around JPEG quality score of 70 (parameter CV\_IMWRITE\_JPEG\_QUALITY defined in \cite{opencv_library}) for both horizontal and oriented models. However, the oriented models had better performances.

Finally a use case is done with a prototype of airplane detection \footnote{\href{https://github.com/satellogic/iquaflow-airport-use-case}{https://github.com/satellogic/iquaflow-airport-use-case}}. In this use case, we apply different compression algorithms on a new dataset. The objective of this experiment is to determine how much each of the compression algorithms affects the performance of an airplane detector. The airplanes dataset consists of  Satellogic images capturing airport areas each of $1024\times1024$ pixels. These captures were made using NewSat Satellogic constellation ($1$~m GSD). The annotations were made using Happyrobot\footnote{\href{https://happyrobot.ai}{https://happyrobot.ai}} platform.

\begin{figure}[ht]
	\centering
	\includegraphics[width=10.5 cm]{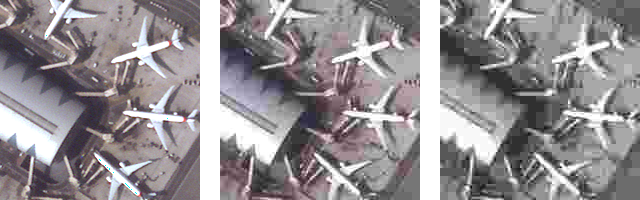}
	\caption{JPEG compression effects (original, JPG10,and JPG5 from left to right).}
	\label{fig:jpeg}
\end{figure}

\begin{figure}[ht]
	\centering
	\includegraphics[width=10.5 cm]{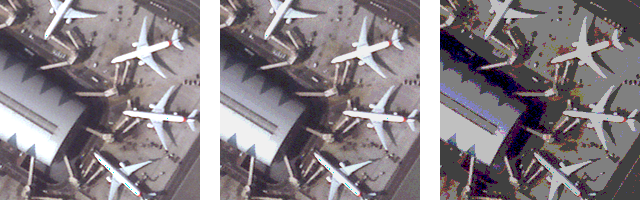}
	\caption{Quantization effects (original, 5 and 2-bits from left to right).}
	\label{fig:quantization}
\end{figure}

\subsection{Super-Resolution} \label{superusecase}

Super-Resolution (SR) refers to any methodology that allows the generation of an image with greater quality which is often represented with a greater amount of pixels. Two use cases were done in this topic: The first one was for Single Frame SR; the other for multi frame SR. They are explained in this section.

The Single Image Super-Resolution (SISR) use case \footnote{\href{https://github.com/satellogic/iquaflow-sisr-use-case}{https://github.com/satellogic/iquaflow-sisr-use-case}} is built to compare the image quality between different SISR solutions. A SiSR algorithm inputs one frame and outputs an image with greater resolution. In this use case we compare the following methods:

\begin{itemize}
\item Fast Super-Resolution Convolutional Neural Network (FSRCNN) \cite{FSRCNN}.
\item Local Implicit Image Function (LIIF) \cite{LIIF}.
\item Multi-scale Residual Network (MSRN) \cite{MSRN}.
\item Enhanced Super-Resolution Generative Adversarial Network (ESRGAN) \cite{ESRGAN}.
\end{itemize}

\begin{figure}[ht][H]
    \centering
    \begin{subfigure}{0.7\textheight}
         \includegraphics[width=\textwidth]{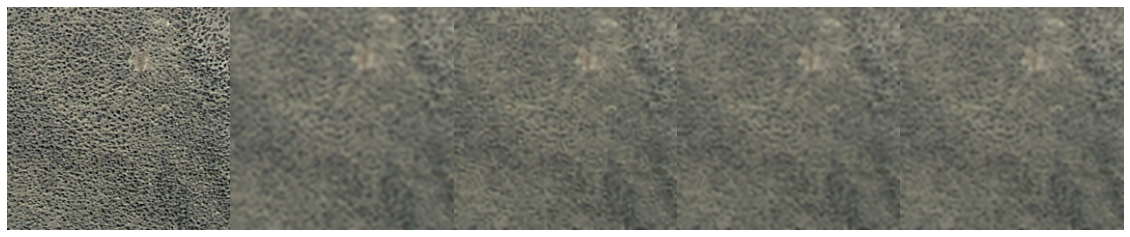}
     \end{subfigure}
     \begin{subfigure}{0.7\textheight}
         \includegraphics[width=\textwidth]{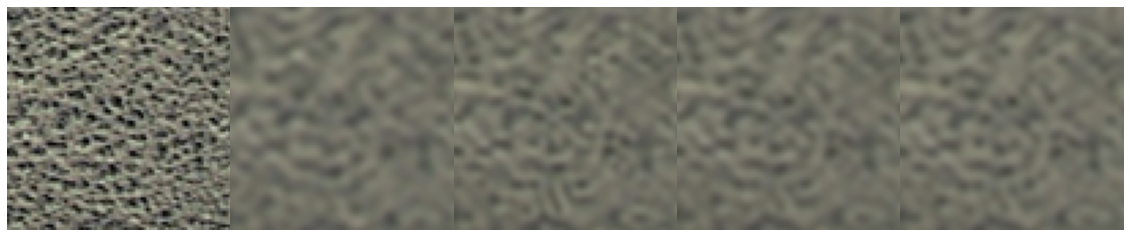}
     \end{subfigure}
     \begin{subfigure}{0.7\textheight}
         \includegraphics[width=\textwidth]{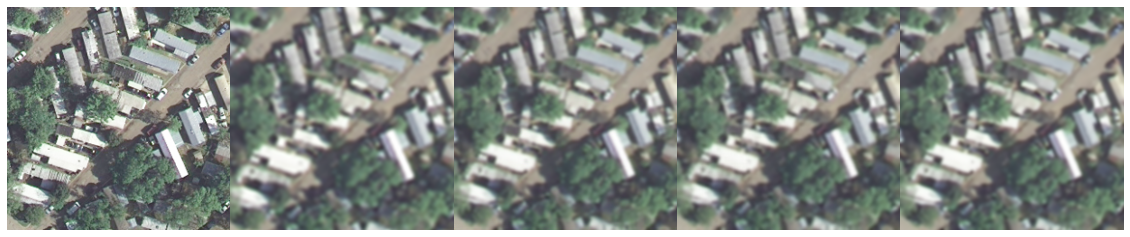}
     \end{subfigure}
     \begin{subfigure}{0.7\textheight}
         \includegraphics[width=\textwidth]{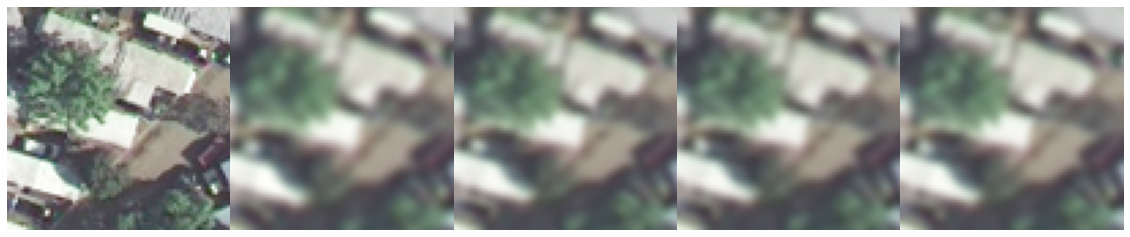}
     \end{subfigure}
\caption{Example of SR solutions. From left to right the images correspond to: The high resolution target image, FSRCNN, LIIF, MSRN and ESRGAN. One can observe how difficult it is to compare quality of solutions that look very similar by human eye. That is one of the reasons of using quantitative methods to assess quality and compare solutions.}
\label{fig:sisrpredictions}
\end{figure}

In this study the public dataset UC Merced Land Use Dataset\cite{UCMerced} was used for training, validating and testing the models. Figure~\ref{fig:sisrpredictions} shows examples of super resolved images from the test partition using each of the methods. The first example on the left belongs to the high resolution ground truth image for comparison. These examples demonstrate that without objective quality metrics it is impossible to evaluate and compare the performance of the different methods. Because of that it is wise to rely on objective image quality tools such as \textsc{Iquaflow}.

\begin{figure}
     \centering
     \begin{subfigure}[b]{0.45\textwidth}
         \centering
         \includegraphics[width=\textwidth]{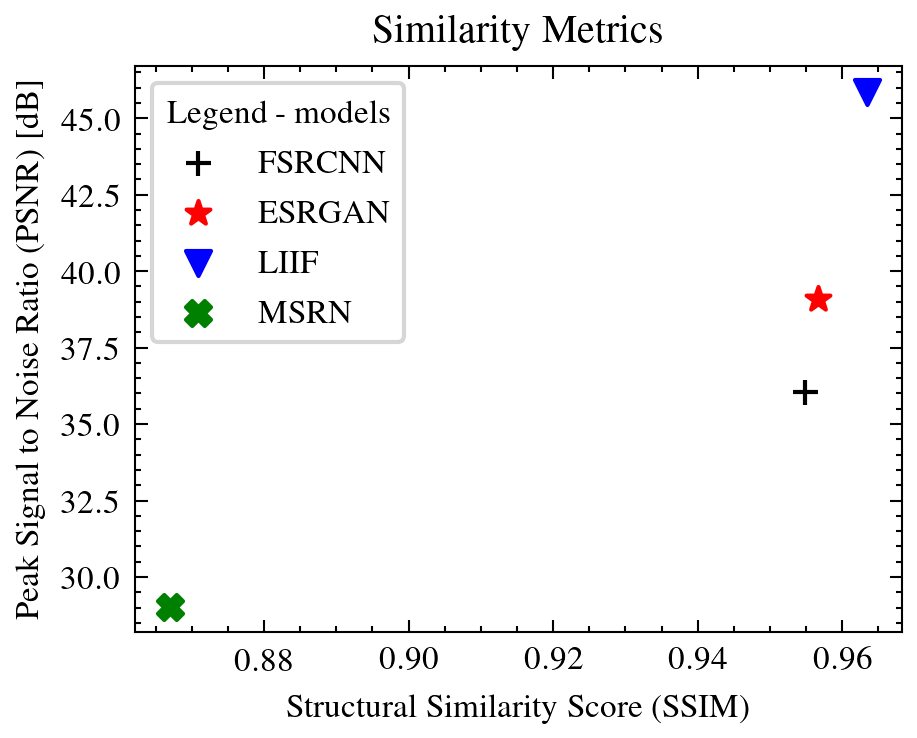}
         \caption{Average Structural Similarity Score (SSIM) and Peak Signal to Noise Ratio (PSNR). These are similarity scores with respect to the target high resolution image.}
         \label{fig:sisrsim}
     \end{subfigure}
     \hfill
     \begin{subfigure}[b]{0.45\textwidth}
         \centering
         \includegraphics[width=\textwidth]{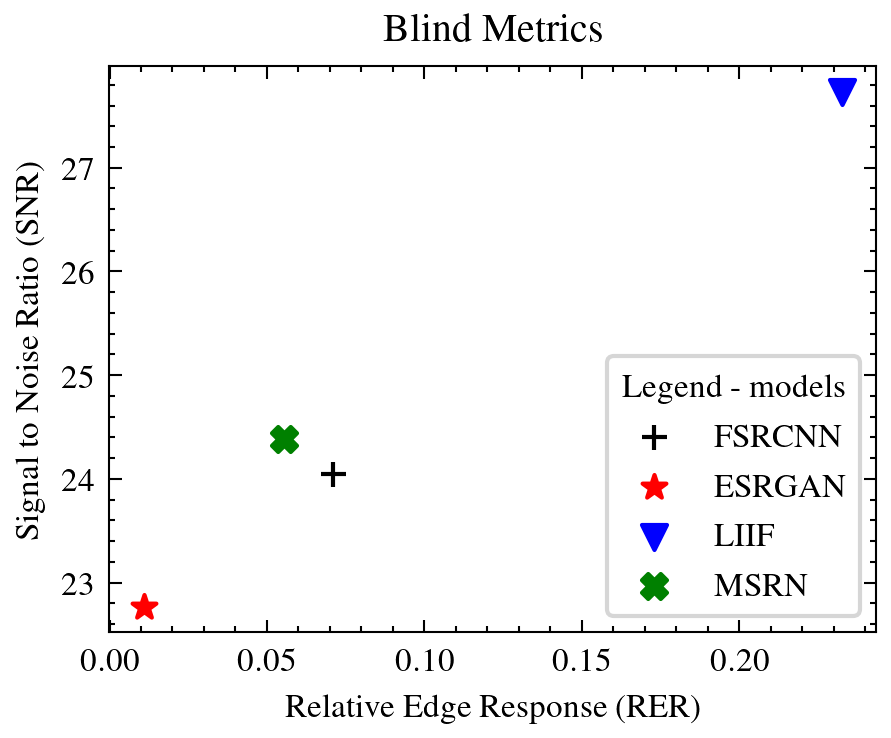}
         \caption{Average Signal to Noise Ratio (SNR) and Edge Response (RER). These are blind metrics that are intrinsic to the image.}
         \label{fig:sisrblind}
     \end{subfigure}
        \caption{Metrics measured on different Single Image Super Resolution (SISR) algorithm solutions}
        \label{fig:sisrmetrics}
\end{figure}

Figure~\ref{fig:sisrmetrics} shows quality metrics measured on the datasets solved by the different solutions of SISR. The graph on the left indicates that the best similarity scores with respect to the target image were achieved with the model LIFF. The graph on the right measures Signal to Noise Ratio (SNR) and Relative Edge Response (RER) which, again, the best values correspond to the LIIF model.

We faced the same challenge of measuring image quality with Multi-Frame Super-Resolution (MFSR) algorithms that are working similarly to the previous algorithms but with a short video (or multiple frames) as inputs rather than a single frame. The MFSR use case\footnote{\href{https://github.com/satellogic/iquaflow-mfsr-use-case}{https://github.com/satellogic/iquaflow-mfsr-use-case}} is build to compare the image quality between different MFSR solutions. A MFSR algorithm inputs several frames of the same scene and outputs a single frame at a greater resolution. In this use case we compare the following methods:

\begin{itemize}
\item Adaptative Gaussian Kernels (AGK) \cite{Wronski_2019}.
\item WarpWeights (WARPW). This is our own method based in geometry and precise location of the multiple frames.
\item HighResNet (HRN) \cite{HighResNet}. HRN is a deeplearning Multiframe SR method. The public dataset xView\cite{xViewdataset} was used for the partitions of training, validation and testing of this model.
\item Multi-scale Residual Network (MSRN) \cite{MSRN}. A single frame algorithm, added as a benchmark for comparison.
\item bi-cubic interpolation. This one is also added as a benchmark for comparison.
\end{itemize}

\begin{figure}
     \centering
     \begin{subfigure}[b]{0.45\textwidth}
         \centering
         \includegraphics[width=\textwidth]{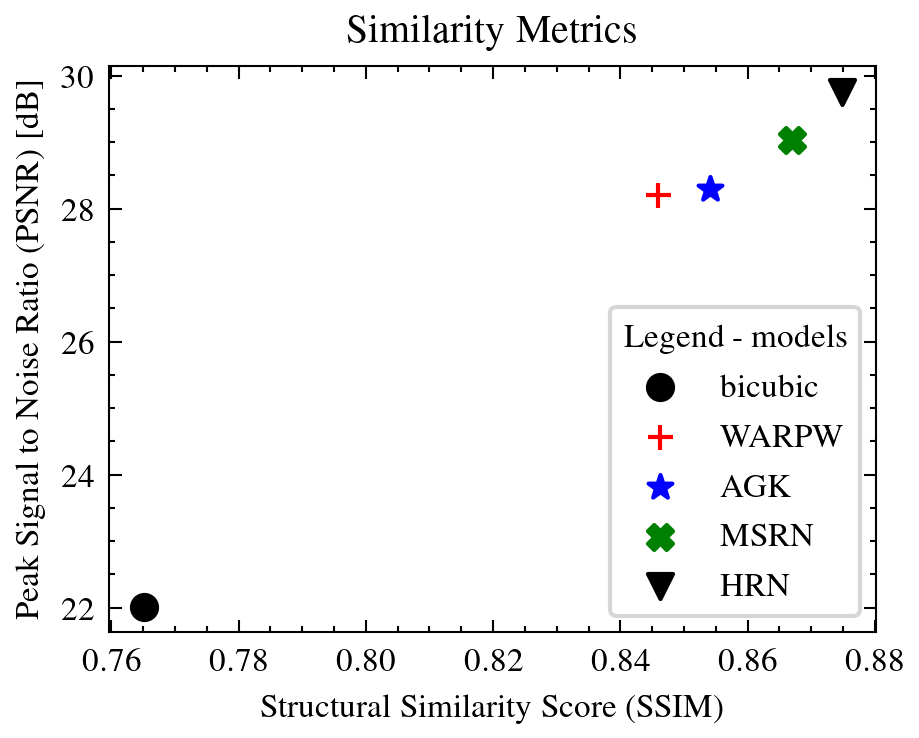}
         \caption{Average Structural Similarity Score (SSIM) and Peak Signal to Noise Ratio (PSNR). These are similarity scores with respect to the target high resolution image.}
         \label{fig:mfsrsim}
     \end{subfigure}
     \hfill
     \begin{subfigure}[b]{0.45\textwidth}
         \centering
         \includegraphics[width=\textwidth]{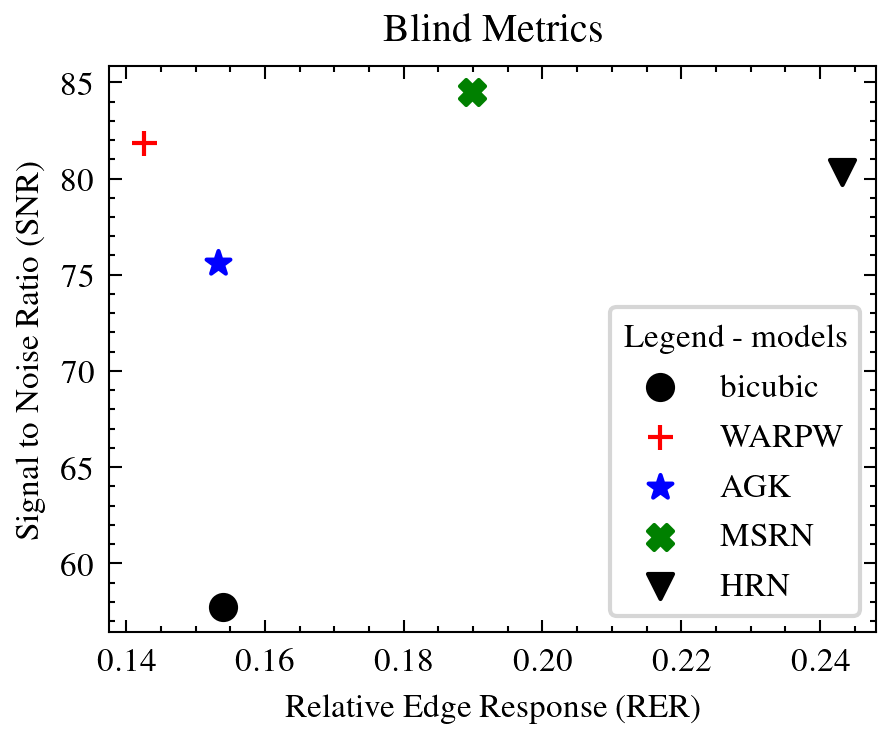}
         \caption{Average Signal to Noise Ratio (SNR) and Edge Response (RER). These are blind metrics that are intrinsic to the image.}
         \label{fig:mfsrblind}
     \end{subfigure}
        \caption{Metrics measured on different Multi Frame Super Resolution (MFSR) algorithm solutions}
        \label{fig:mfsr}
\end{figure}

In this case MSRN corresponds to a Single Frame algorithm and it is included as a benchmark reference of a single frame algorithm to compare with. Figure~\ref{fig:mfsr} shows the quality metrics measured in the resulting images of the different methods. The similarity scores are indicating a clear improvement with respect to a simple bi-cubic interpolation. The best methods seem to be the ones based on deep learning (HRN and MSRN). Furthermore, Multiframe deep learning solution has greater score than MSRN. Despite the lower scores of methods AGK and WARPW they also are executed much faster and with less computational resources.

\section{Conventions} \label{Conventions}

Many tools and frameworks are designed by the philosophy of “convention over configuration” allowing the following small set of conventions to work in an easy manner with the tool. With this approach \textsc{iquaflow} can be adapted to any kind of deep learning model and custom training loop. Thus, we defined a set of conventions that the user must adopt in order to create a correct \textsc{iquaflow} analysis.

\subsection{Dataset Formats} \label{sectiondatasetformats}

\begin{figure}[ht]
	\centering
	\includegraphics[width=15 cm]{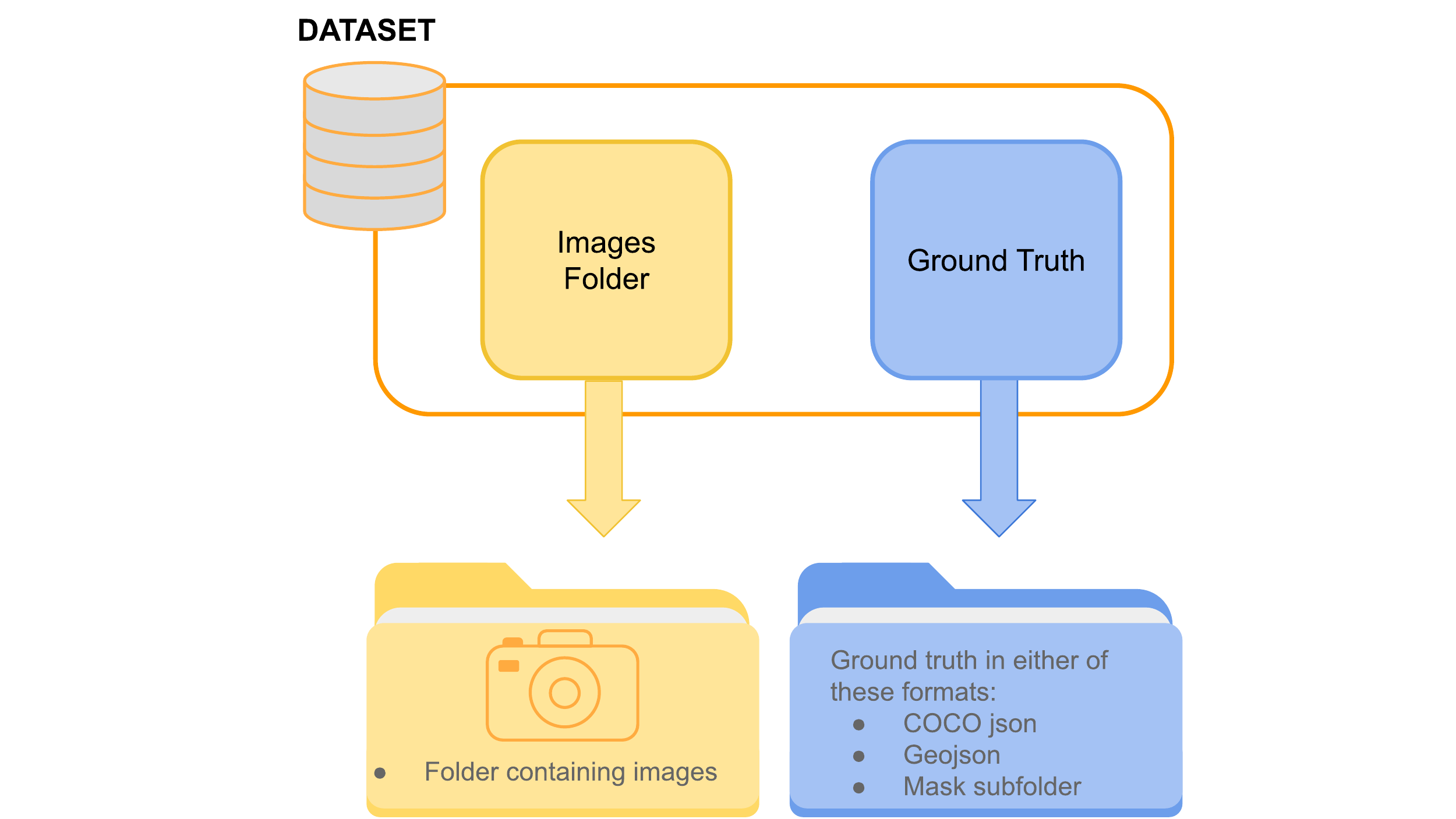}
	\caption{Dataset structure in \textsc{iquaflow}.}
	\label{fig:dsstruct}
\end{figure}

\textsc{iquaflow} understands a dataset as a folder containing a sub-folder with images and ground truth (See Figure~\ref{fig:dsstruct}). The name of the sub-folder can be anything that does not contain the word "mask" which is reserved for mask annotations. Datasets that do not follow this format should be adapted in order to perform experiments. In case of detection or segmentation tasks any of these options are supported formats:
\begin{itemize}
\item Json in COCO \cite{lin2014microsoft} format.
\item GeoJson with the minimum required fields: image\_filename, class\_id and geometry.
\item A folder named masks with images corresponding to the segmentation annotations.
\end{itemize}

\begin{figure}[ht]
	\centering
	\includegraphics[width=15 cm]{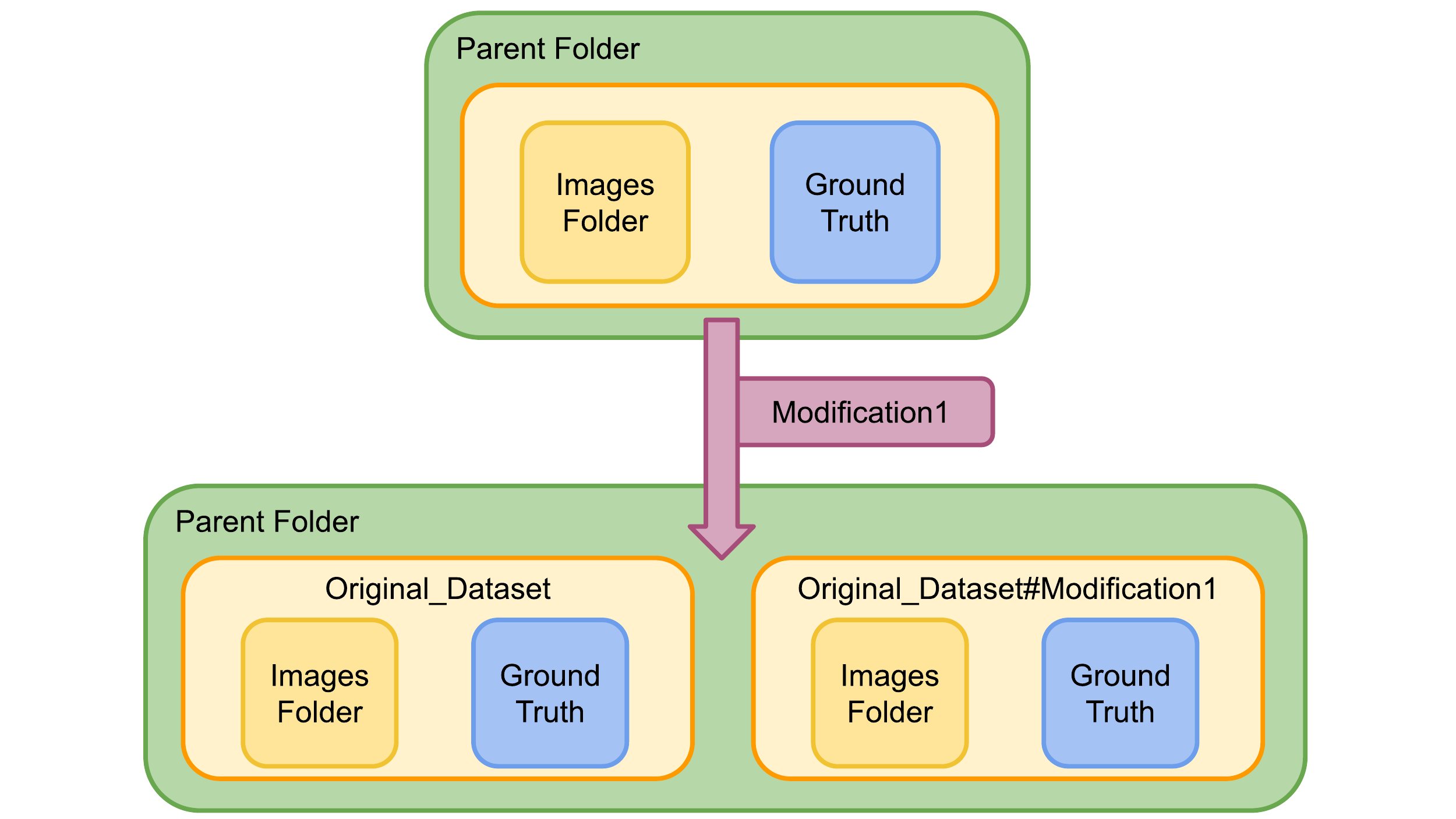}
	\caption{Original and modified datasets.}
	\label{fig:modds}
\end{figure}

\textsc{iquaflow} primarily works with COCO \cite{lin2014microsoft} json ground truth adopted by most of the datasets and models of the field. In case that the dataset is in other format, the user can transform it to COCO \cite{lin2014microsoft}. Otherwise, \textsc{iquaflow} neither can perform sanity nor statistics checks. For other kind of tasks, such as image generation, it is only necessary to have the ground truth in a json format. Alternatively, \textsc{iquaflow} can recognize a dataset without any ground truth file. When the dataset is modified, \textsc{iquaflow} creates a modified copy of the dataset in its parent folder. As a convention, \textsc{iquaflow} adds to the name of the original dataset a “\#” followed by the name of the modification as seen in Figure~\ref{fig:modds}.

\subsection{Training script} \label{trainingscriptsubsection}

\begin{figure}[ht]
 	\centering
 	\includegraphics[width=15 cm]{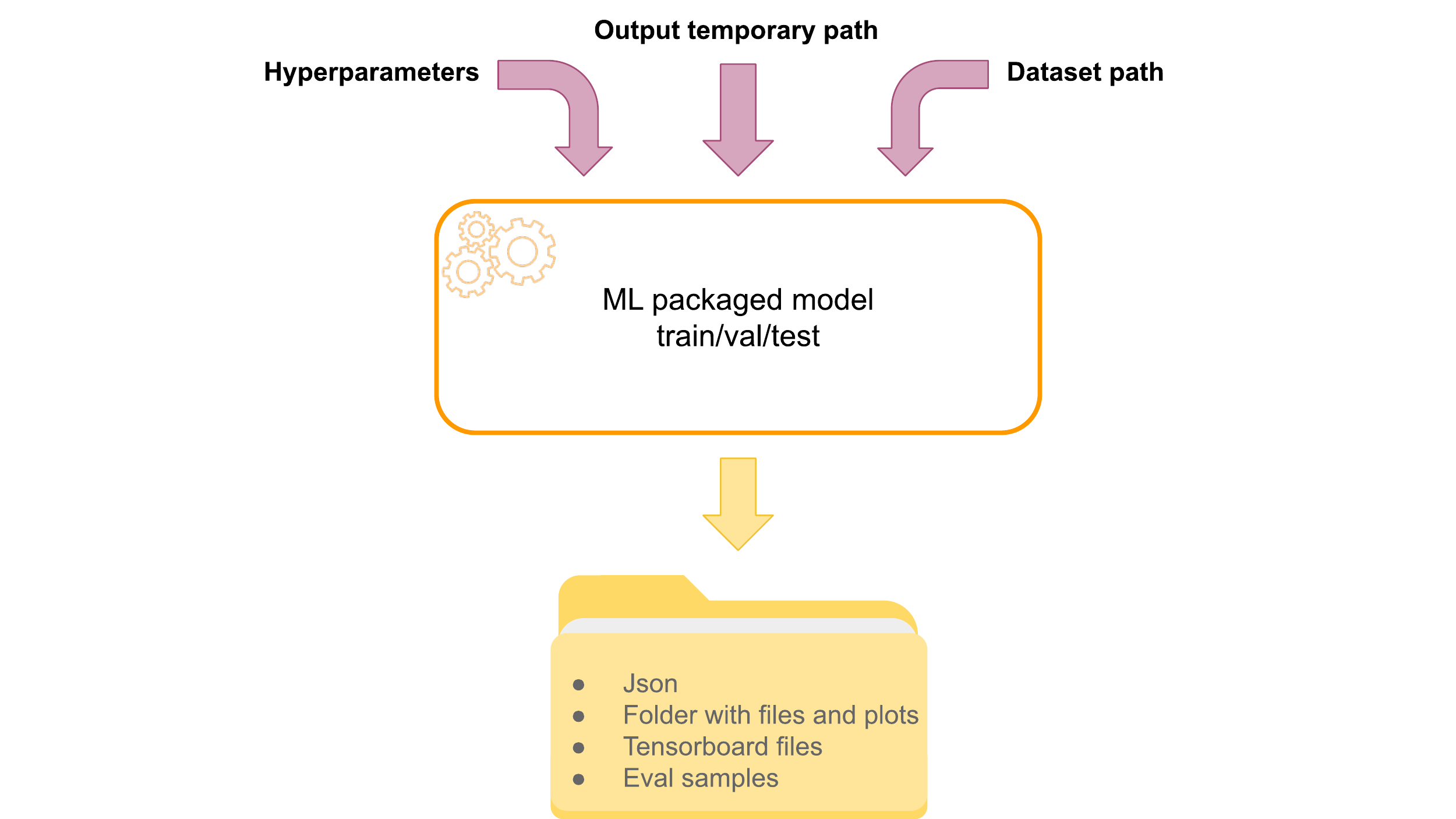}
 	\caption{Training script example structure.}
 	\label{fig:trainingscript}
\end{figure}

The user is responsible to provide the model training scripts containing the custom training loop of the model. In this way \textsc{iquaflow} is agnostic to any kind of model or deep learning framework, it interacts with deep learning as a black box, as you can see in the Figure~\ref{fig:trainingscript}. 

There are only a few input arguments that are necessary and allow the interaction of \textsc{iquaflow} with the model. Thus, the training script requires the following arguments by convention:

\begin{itemize}
\item \textbf{outputpath}:  \textsc{iquaflow} logs all the content of this path as artifacts in \textsc{Mlflow}. Furthermore, it will log as metrics and parameters the files that are following the convention of the outputs as stated in \ref{outputformats}.
\item \textbf{trainds}:  The training dataset path.
\item \textbf{valds (optional)}:  The validation dataset path.
\item \textbf{testds (optional)}:  The testing dataset path.
\item \textbf{mlfuri (optional)}:  The url pointing to the \textsc{Mlflow} server.
\item \textbf{mlfexpid (optional)}:  The experiment id.
\item \textbf{mlfrunid (optional)}:  The run id.
\item \textbf{other hyperparameters (optional)}:  Any other custom parameter that belongs to the user training script. These parameters can be specified afterwards in the ExperimentSetup of \textsc{iquaflow}. They can also be varied by specifying a list with the different values.
\end{itemize}

Note that these arguments are fed to the user's training script by \textsc{iquaflow} itself as the framework is responsible for combining the parameters into a set of runs, executing modifications of the dataset and logging the details of the experiments in \textsc{Mlflow}. Optional parameters are only added when they are also specified in the ExperimentSetup. It is the case of the arguments starting with \texttt{mlf} which are only used when the flag mlflow\_monitoring is activated in the ExperimentSetup. Their purpose is to monitor in streaming the training, without them the user will still get everything logged in \textsc{Mlflow} at the end of each training.

\subsection{Output Formats} \label{outputformats}

Any of the data that is saved by the user training script in the output folder will be picked up by \textsc{iquaflow} and be parsed as experiment parameters, metrics or artifacts. These are the key files that can be saved by convention in the output folder:

\begin{itemize}
\item \textbf{results.json}: JSON with the names of parameters and metrics for the keys. Then the values are a single scalar for parameters and a list of values for metrics.

\begin{mdframed}
\begin{minted}{python}
    {
     "learning_rate": 0.83,
     "num_epochs": 100,
     "train_focal_loss": [1.34, 1.29, 1.24, 0.01]
     "val_focal_loss": [1.34, 1.29, 1.24, 0.01]
    }
\end{minted}
\end{mdframed}

\item \textbf{output.json}: It is the predicted output in COCO json format\cite{lin2014microsoft}. It contains as many elements as detections have been made in the dataset. An object detection metric will rely on this file.

\begin{mdframed}
\begin{minted}{python}
    {
        "image_id" : 85
        "iscrowd" : 0
        "bbox":[
            522.5372924804688
            474.1499938964844
            28.968505859375
            27.19696044921875
        ]
        "area": 2427.050960971974
        "category_id": 1
        "id": 1
        score : 0.9709288477897644
    }
        \end{minted}
\end{mdframed}

\item \textbf{Image generation}: The json may contain the relative path to the generated images. Imagine the packaged model is Super Resolution model that generates five super resolution images. The package may store a folder named generated\_sr\_image in the output temporary file with this five images. Hence the output.json should be as following:

\begin{mdframed}
\begin{minted}{python}
    {
     [
       "generated_sr_image/image_1.png",
       "generated_sr_image/image_2.png",
       "generated_sr_image/image_3.png",
       "generated_sr_image/image_4.png",
       "generated_sr_image/image_5.png",
     ]
    }
        \end{minted}
\end{mdframed}

\end{itemize}

\section{Pre-Processing Stage}

SanityCheck and DSStatistics are the classes that will perform sanity check and statistics of image datasets and ground truth. They are stand alone classes, it is to say they can work by proving the path folder of images and ground truth, or they can work with DSWrapper class.

\subsection{Sanity check}

The SanityCheck module performs sanity to image datasets and ground truth. It can either work as standalone class or with DSWrapper class. It will remove all corrupted samples following the logic in the argument flags. The new sanitized dataset is located in output\_path attribute from the SanityCheck instance. A usage example:

\begin{mdframed}
\begin{minted}{python}
from iquaflow.sanity import SanityCheck

sc = SanityCheck(data_path, output_folder)
sc.check_annotations()
    \end{minted}
\end{mdframed}

These are the tasks that are done when calling the check\_annotations method:
\begin{itemize}
\item Finding duplicates in COCO \cite{lin2014microsoft} json images list.
\item Check if the image format is a valid image file format.
\item Check integrity of one COCO \cite{lin2014microsoft} annotation.
\item Fix height and width in COCO \cite{lin2014microsoft} json images list.
\item In geojson annotations, remove all rows containing a Nan value, empty geometries in any of the required field columns.
\item In geojson annotations, try to fix geometries with buffer = 0 and remove the persistent invalid geometries.
\end{itemize}

Note the difference between missing, empty and invalid geometries in a geojson:

\begin{itemize}
\item \textbf{Missing geometries}: This is when the attribute geometry is empty or unknown. Most libraries load it as None type in python. These values were typically propagated in operations (for example in calculations of the area or of the intersection), or ignored in reductions such as unary\_union.
\item \textbf{Empty geometries}: This happens when the coordinates are empty despite having a geometry type defined. This can happen as a result of an intersection between two polygons that have no overlap.
\item \textbf{Invalid geometries}: Problematic features such as edges of a polygon intersecting themselves. This could have happened due to a mistake from the annotator. For the case of invalid geometry. The tool will also attempt to fix them with buffer=0 functionality prior to removing. In future releases an additional argument to simplify geometries will be offered.
\end{itemize}

\subsection{Statistics and exploration}

The module DsStats calculates different statistics on the image datasets and their annotations. It allows the user to explore the data interactively, and to visualize the statistics, as well as to export them for later use. It can either work as standalone class or with DSWrapper class. A usage example:

\begin{mdframed}
\begin{minted}{python}
from iquaflow.ds_stats import DsStats

dss = DsStats(data_path, output_folder)
stats = dss.perform_stats(show_plots = True)
    \end{minted}
\end{mdframed}

Calling the pserform\_stats method calculates and reports the following statistics:
\begin{itemize}
\item Average height and width of the images.
\item Histogram of the occurrences of the classes in the data.
\item Image and bounding box aspect ratio and area histograms.
\item Auto-generates the best fitting bounding box and rotated bounding box of the dataset annotations. It also adds the high, width angle from this new generated attributes.
\item Compactness, centroid and area of the polygons.
\item Estimates the min, mean and max from any given dataframe field from annotations that are in geojson format.
\end{itemize}

There are also two interactive exploratory tools. Some functionalities can be used in line in notebooks or exported as an interactive html. The first tool is for visualizing the annotations an the second tool shows a summary of some images. These are:
\begin{itemize}
\item notebook\_annots\_summary.
\item notebook\_imgs\_preview.
\end{itemize}

Usage example:

\begin{mdframed}
\begin{minted}{python}
from iquaflow.ds_stats import DsStats

DsStats.notebook_annots_summary(
    df,
    export_html_filename=html_filename,
    fields_to_include=["image_filename", "class_id", "area"],
    show_inline=True,
)

from iquaflow.ds_stats import DsStats

DsStats.notebook_imgs_preview(
        data_path=data_path,
        sample=100,
        size=100,
)
    \end{minted}
\end{mdframed}

\subsection{Dataset}

DSWrapper is the class that \textsc{iquaflow} uses for handling datasets. Section~\ref{sectiondatasetformats} explains the conventions of a dataset in \textsc{iquaflow}. To create an instance of DSWrapper one must specify the location path to the dataset as follows:

\begin{mdframed}
\begin{minted}{python}
from iquaflow.datasets import DSWrapper
ds_wrapper = DSWrapper(data_path="[path_to_the_dataset]")
    \end{minted}
\end{mdframed}

\textsc{iquaflow} parses the dataset structure the following way:

\begin{itemize}
\item \textbf{ds\_wrapper.parent\_folder}: Path of the folder containing the dataset.
\item \textbf{ds\_wrapper.data\_path}: Root path of the dataset.
\item \textbf{ds\_wrapper.data\_input}: Path of the folder that contains the images.
\item \textbf{ds\_wrapper.json\_annotations}: Path to the json annotations. Preferred COCO \cite{lin2014microsoft} annotations.
\item \textbf{ds\_wrapper.geojson\_annotations}: Path to the geojson annotations.
\item \textbf{ds\_wrapper.params}: Contains metainfomation of the dataset. Initially {"ds\_name":"[name\_of\_the\_dataset]"}
\end{itemize}

Furthermore, DSWrapper contains an editable dictionary that describes the dataset. Initially this dictionary contains the key ds\_name that is the name of the dataset. The user can populate this dictionary with any key/value parameter. Afterwards, this dictionary will be populated and changed automatically by DSModifier classes and it will be used for experiments logs.

\subsection{Modifiers}

Modifiers take a dataset $D$ and process it to obtain a $D^\prime$ dataset with the modification defined by the user. There are built-in modifiers and the user can also create their own. When the dataset is modified, \textsc{iquaflow} creates a modified copy of the dataset in its parent folder. As a convention, \textsc{iquaflow} adds to the name of the original dataset a "\#" followed by the name of the modification as you can see in the following image. To use a modifier one just needs to import the desired modifier and run it.

\begin{mdframed}
\begin{minted}{python}
from iquaflow.datasets import DSModifier_jpg

img_path = "test_datasets/ds_coco_dataset/images")
jpg85 = DSModifier_jpg(params={"quality": 85})
jpg85.modify(data_input=img_path)
    \end{minted}
\end{mdframed}

After running, a \emph{test\_datasets/ds\_coco\_dataset\#jpg85\_modifier/images/} folder should be created with the modified images.

Alternatively any new designed custom modifier can be integrated in \textsc{iquaflow}. To do so one can follow the example in the code module \emph{modifier\_jpg.py}. The user needs to inherit from \emph{DSModifier\_dir} and write the \emph{internal\_mod\_img()} member function.

\section{Experiment}

\subsection{Task Execution}

\textsc{iquaflow} can automatize and log all the information that the user saves in the output files folder. This way, the user has the flexibility to log experiment information without knowing any specific logging tool. TaskExecution is the generic class that wraps the user packaged model and provides arguments to the training script. It is also responsible for translating all the experiment information to the \textsc{Mlflow} tracking server. Ultimately, \textsc{iquaflow} uses \textsc{Mlflow} to organize the experiments and the user does not need to understand how \textsc{Mlflow} works.

\subsubsection{PythonScriptTaskExecution}

This particular class extends from TaskExecution and knows how to execute a model that is encapsulated in a python script. The user needs to instantiate the class, adding the path to the python script as argument.

\begin{mdframed}
\begin{minted}{python}
task = PythonScriptTaskExecution(model_script_path="./path_to_script.py")
    \end{minted}
\end{mdframed}

Alternatively the user can execute the task, but is not recommended since \textsc{iquaflow} will perform executions internally when the whole experiment is defined. In order to execute the run, the user must provide the experiment name, the name of the run and the training dataset path or training DSWrapper. Optionally, the user can provide a training dataset path or ds\_wrapper and a python dictionary with model hyper\_parameters (that will be used when executing the package)

\begin{mdframed}
\begin{minted}{python}
task.train_val(
            experiment_name="name of the experiment",
            run_name="test_run",
            train_ds=ds_wrapper_train,
            val_ds=ds_wrapper_validation,
            hyper_parameters={"lr": 1e-6},
        )
\end{minted}
\end{mdframed}

\subsubsection{SagemakerTaskExecution}

Our application can run on \href{Amazon Sagemaker}{https://aws.amazon.com/sagemaker/} by passing a SageMakerEstimatorFactory as an argument of our TaskExecution. In which case it becomes a SageMakerTaskExecution. See an example on how to define it.

\begin{mdframed}
    \begin{minted}{python}
    from sagemaker.pytorch import PyTorch
    from iquaflow.experiments.task_execution import SageMakerEstimatorFactory, SageMakerTaskExecution
    
    sage_estimator_factory = SageMakerEstimatorFactory(
       PyTorch,
       {
           "entry_point": "train.py",
           "source_dir": "yolov5",
           "role":role,
           "framework_version": "1.8.1",
           "py_version": "py3",
           "instance_count": 1,
           "instance_type": "ml.g4dn.xlarge"
       }
    )
    
    task = SageMakerTaskExecution( sage_estimator_factory )
    \end{minted}
\end{mdframed}

Then in the user training script, one might want to connect the argument script variables that are defined by convention in \textsc{iquaflow} (see Conventions \ref{Conventions}) to SageMaker environmental variables to take full advantage of the SageMaker tools. As an example:

\begin{mdframed}
\begin{minted}[fontsize=\scriptsize]{python}
import argparse

parser = argparse.ArgumentParser()

# Define some defaults
trainds_default     = (os.environ["SM_CHANNEL_TRAINDS"] if "SM_CHANNEL_TRAINDS" in os.environ else "")
valds_default      = (os.environ["SM_CHANNEL_VALDS"] if "SM_CHANNEL_VALDS" in os.environ else "")
outputpath_default = (os.environ["SM_OUTPUT_DATA_DIR"] if "SM_OUTPUT_DATA_DIR" in os.environ else "./output")

# IQF arguments
parser.add_argument("--trainds", default=trainds_default, type=str, help="training dataset path")
parser.add_argument("--valds", default=valds_default, type=str, help="validation dataset path")
parser.add_argument("--outputpath", default=outputpath_default, type=str, help="path output")
    \end{minted}
\end{mdframed}

Also, for these approaches one might want \textsc{iquaflow} to upload the modifed datasets (by \textsc{iquaflow}-modifiers) on a bucket on the fly. To do so, the user must indicate the bucket\_name in the cloud\_options whithin ExperimentSetup.

\subsubsection{Experiment Setup}

\textsc{iquaflow} allows to formulate experiments taking as reference the modified training dataset. In order to perform this task, the package provides tools that allow to automatize these kinds of experiments that are composed by:

\begin{itemize}
\item A training dataset.
\item A list of dataset modifiers.
\item A machine learning Task.
\end{itemize}

The first two components are covered by DSWrapper and DSModifer respectively. The last one requires a TaskExecution. Having defined all the components the user is able to perform a \textsc{iquaflow} experiment by using ExperimentSetup. The user must define the name of the experiment, the reference datasets, the list of datasets modifiers and the packaged model, as following

\begin{mdframed}
\begin{minted}{python}
experiment = ExperimentSetup(
   experiment_name="experimentA",
   task_instance=PythonScriptTaskExecution(model_script_path="./path_to_script.py"),
   ref_dsw_train=DSWrapper(data_path="path_to_dataset"),
   ds_modifiers_list=[ DSModifier_jpg(params={'quality': i}) for i in [10,30,50,70,90] ]
)
    \end{minted}
\end{mdframed}

And then just execute the experiment by:

\begin{mdframed}
\begin{minted}{python}
experiment.execute()
    \end{minted}
\end{mdframed}

Some other options for the ExperimentSetup are:

\begin{itemize}
\item \textbf{repetitions}: Each combination of parameters and modifiers results in a run. Scipts might contain randomness (i.e. Random partitions). For those cases the user might want to average out several executions to have a relevant statistic or study the variability. To do so, one can set the number of repetitions to greater than 1.
\item \textbf{mlflow\_monitoring}: This allows monitoring the training script in real time. When turned on, \textsc{iquaflow} will pass these additional arguments to the training script (See section \ref{trainingscriptsubsection}):
    \begin{itemize}
        \item \textbf{mlfuri}
        \item \textbf{mlfexpid}
        \item \textbf{mlfrunid}
    \end{itemize}
Thus, the user will be responsible to add these in the user training script when required. Then the user can activate the current experiment and run in the the script with a snippet such as:

\begin{mdframed}
\begin{minted}{python}
mlflow.set_tracking_uri(args.mlfuri)

mlflow.start_run(
    run_id=args.mlfrunid,
    experiment_id=args.mlfexpid
)
    \end{minted}
\end{mdframed}

\item \textbf{cloud options}: It is a dictionary of options useful for indicating endpoints such as:
    \begin{itemize}
    \item \textbf{bucket\_name}: If set, modified data (by \textsc{iquaflow}-modifiers) will be uploaded to the bucket.
    \item \textbf{tracking\_uri}: trackingURI for \textsc{Mlflow}. default is local to the ./mlflow folder
    \item \textbf{registry\_uri}: registryURI for \textsc{Mlflow}. default is local to the ./mlflow folder
    \end{itemize}

Indicating the bucket is useful for SageMakerTaskExecution instances.
\end{itemize}

\section{Managing Results}

\subsection{Experiment Info}

This class allows to manage the experiment information. It simplifies the access to \textsc{Mlflow} and allows to apply new metrics to previous executed experiments. Basic usage example:

\begin{mdframed}
\begin{minted}{python}
from iquaflow.experiments import ExperimentInfo

experiment_info = ExperimentInfo(experiment_name)
runs = experiment_info.get_mlflow_run_info() # runs is a python dict
    \end{minted}
\end{mdframed}

These are the main methods:

\begin{itemize}
\item \textbf{get\_mlflow\_run\_info}: It gathers the experiment information in a python dictionary.
\item \textbf{apply\_metric\_per\_run}: Applies a new metric to previously executed experiments.
\item \textbf{get\_df}: Retrieves a selection of data in a suitable format so that it can be used as an input in the Visualization module\ref{visualizationsection}.
\end{itemize}

In the following sections, we provide some examples of how to use the last two methods.

\subsection{Metrics}

The module metrics contains functionalities to estimate model performance metrics. \textsc{iquaflow} includes some metrics and it also provides a generic Metric class that allows the user to easily implement their own custom metrics. To make a custom metrics one must inherit from the class Metrics as follows:

\begin{mdframed}
\begin{minted}{python}
    from iquaflow.metrics import Metric
    
    class CustomMetric(Metric):
        def __init__(self) -> None:
            self.metric_names = coco_eval_metrics_names
        def apply(self, predictions: str, gt_path: str) -> Any:
            # Your custom code here
            # Then return a dictionary of names and values for each metric
            return {k: v for k, v in zip(metric_names, stats)}
        \end{minted}
\end{mdframed}

Then to calculate a metric to an executed experiment do:

\begin{mdframed}
\begin{minted}{python}
    from iquaflow.experiments import ExperimentInfo
    
    experiment_info = ExperimentInfo(experiment_name)
    my_custom_metric = CustomMetric()
    experiment_info.apply_metric_per_run( my_custom_metric, json_annotations_name )
\end{minted}
\end{mdframed}

Some relevant available metrics offered by \textsc{iquaflow} are described in the following sections:

\subsection{BBDetectionMetrics}

BBDetectionMetrics is a metric for object detection. It can be applied between bounding boxes of ground truth and predicted elements. The the ground-truth annotations must be in COCO format and the predictions in COCO-inference format\cite{lin2014microsoft} ( See COCO detection and COCO data ). When this metric is applied the metrics from COCOeval (See COCO detection ) are estimated. This includes metrics such as (Recall, mAP, etc.)
\begin{mdframed}
\begin{minted}{python}
    from iquaflow.metrics import BBDetectionMetrics
        \end{minted}
\end{mdframed}

\subsection{SNRMetric}

Signal-to-noise ratio (SNR) is a metric that measures the strength/level of the considered signal relative to the background noise. Currently there are two algorithms implemented in \textsc{iquaflow} that measure SNR of an image: homogeneous blocks (HB; the default method) and homogeneous area (HA). The main difference between the two methods is that while HB uses the whole image, HA intends to find homogeneous areas in the image to calculate the SNR. This leads to a trade-off: HB is faster than HA; HB returns a value every time while HA return a None value if it fails to find homogeneous areas; HB has higher uncertainty than HA. Both methods result in approximated values only, since a precise measurement requires a more careful consideration of the image content.


\begin{mdframed}
\begin{minted}{python}
from iquaflow.metrics import (
       SNRMetric,
       snr_function_from_array,
       snr_function_from_fn
)
    \end{minted}
\end{mdframed}

\subsection{SharpnessMetric}

Image sharpness can be defined in several ways. The most common metrics include the relative edge response (RER), the full-width half maximum (FWHM) value of the point spread function (PSF) of the image, or the modulation transfer function (MTF), which is often evaluated the Nyquist frequency \citep{sharpness}. The SharpnessMetric class implements all three metrics, and reports the values in 3 edge directions. The implemented algorithms are based on \cite{cenci2021presenting}.


\begin{mdframed}
\begin{minted}{python}
from iquaflow.metrics import SharpnessMetric
    \end{minted}
\end{mdframed}



\section{Visualization} \label{visualizationsection}

Apart from the visualization tools explained in the Sanity check and Statistics section, there are also tools for plotting and summarizing the results. One key service is \textsc{Mlflow} that is accessed from the browser and allows to visualize, query and compare runs. These are the main features of \textsc{Mlflow}:

\begin{itemize}
\item Experiment-based run listing and comparison.
\item Searching for runs by parameter or metric value.
\item Visualizing run metrics.
\item Downloading run results.
\end{itemize}

Furthermore the ExperimentVisual class offers plotting utilities that can be either inline or saved into files. It uses a pandas data-frame\cite{reback2020pandas} extracted from an ExperimentInfo as input.

\section{Contributing}

These are the tools, environment and procedures required for developing and collaborating with this project.

\subsection{Package Overview}

The python package structure of this tool box is based on \href{https://github.com/cookiecutter/cookiecutter}{cookiecutter}. This library provides a standard workflow for developing production level packages. The tools that will be used are: 

\begin{itemize}
\item \textbf{setuptools} for packaging 
\item \textbf{versioneer} for versioning 
\item \textbf{GitLab CI} for continuous integration 
\item \textbf{tox} for managing test environments 
\item \textbf{pytest} for tests 
\item \textbf{sphinx} for documentation 
\item \textbf{black}, \textbf{flake8} and \textbf{isort} for style checks 
\item \textbf{mypy} for type checks
\end{itemize}

More information can be found in: 
\begin{itemize}
\item \href{https://packaging.python.org/tutorials/packaging-projects/}{python\_packaging}.
\item \href{https://python-packaging.readthedocs.io/en/latest/minimal.html}{readthedocs}.
\item \href{https://www.learnpython.org/en/Modules_and_Packages}{Modules\_and\_Packages}.
\end{itemize}

\subsection{Environment installation}

This repository does not require any specific python environment. The file setup.py allows to install \textsc{iquaflow} as a python package via pip. Once the new environment has been created, one must clone the repository. Then the user can do the wallowing command to install the \textsc{iquaflow} as a softlink in the environment. Dependencies are defined in setup.cfg under install\_requires tag. So first the package is installed in the local environment and then the dependency is added in the setup.cfg with its corresponding version.

\subsection{Documentation}

We use Sphinx to automatically update our documentation. This allows to update package documentation as new code is added. The documentation and Sphinx configuration can be found inside /doc.

\subsection{Continuous integration}

In our project we use \href{https://tox.wiki/en/latest/index.html}{TOX}. This tool allows to manage multiple environments in order to automatically validate code. More information about TOX can be found in \href{https://tox.wiki/en/latest/index.html}{here}.

\begin{mdframed}
\begin{minted}{bash}
# For quality check:
tox -e check

# For automatic code reformat:
tox -e reformat

# For executing all test for first time use
tox -r -e py36

# Alternatively, if it is not the first time it is not necessary to recreate the tox environment.
tox -e py36
    \end{minted}
\end{mdframed}

\subsection{Testing}

Unit tests are performed using PyTest. All tests are included in the test folder located in the repository main folder. Once a new test module that includes python assertions is made ( e.g. test\_new\_module). Then one must simply type in the console pytest or:

\begin{mdframed}
\begin{minted}{bash}
pytest <module name>
    \end{minted}
\end{mdframed}

to run the tests.

We strongly recommend to use “test\_” as the prefix of every test you create.

One can also run test manually using tox(recommended) (use -r parameter for creating tox environment for the first time):

\begin{mdframed}
\begin{minted}{bash}
tox -e py36
    \end{minted}
\end{mdframed}

More information can be found in \href{here.https://docs.python-guide.org/writing/tests/}{here}.

\subsection{Initial development process}

Below we describe usual steps when developing from scratch:

\begin{mdframed}
\begin{minted}{bash}
# Setup python environment:
conda create -n iqt-env python=3.6
# Clone repository:
git clone https://github.com/satellogic/iquaflow
#Create branch:
git checkout -b <new_branch_name>
# Install soft link via:
python -m pip install -e .
# Create test that defines modules functionality.
# Solve the test by adding package functionality.
# If new branch pulled use tox -r to recreate tox environments.
# Reformat code:
python -m pip install tox
tox -e reformat
# Check code and solve:
tox -e check
# Run tests:
tox -e py36
# Push to remote branch.
# Create MR and assign reviewer.
# Refreshing local repository for running tests (after pip install -e .):
tox -r -e py36Sphinx
    \end{minted}
\end{mdframed}

\section{Acknowledgements}

\subsection{Sponsor}

The project was financed by the Ministry of Science and Innovation and by the European Union within the framework of Retos-Colaboraci{\'o}n of the Research, Development and Innovation State Program Oriented to the Challenges of Society, within the Scientific, Technical and Innovation State Research Plan 2017-2020, with the main objective of promoting technological development, innovation and quality research. Project reference >> RTC2019-007434-7.

\subsection{Development}

The project has been coordinated by SATELLOGIC and with the participation of the Multimedia Technologies Unit of Eurecat along with the Group on Interactive Coding of Images (GICI) at Universitat Aut{\`o}noma de Barcelona (UAB). It is a multidisciplinary team, which was absolutely necessary to successfully achieve the objectives set out in the project. SATELLOGIC, as project leader, assumed a decisive role in the management and coordination, as well as in the analysis of specifications and requirements. Its experience in the development of solutions applicable to the field of OE systems allowed it to lead the design and development of the \textsc{iquaflow} Framework, acting as responsible for the technical coordination of the project. It also participated in the integration of the final prototype, validation tests, functional tests and the integration and validation of the complete system both at laboratory level and at real environment validation level. UAB-DEIC-GICI took advantage of its experience in the field of data compression, in particular in remote sensing image coding on board satellite to work in the design, development and characterization of a Compression System. They also played an active role in the Definition and Scoping of the project as well as the final prototype and validation tests for the code and configurations of the compression system software implementation. EURECAT performed the design and implementation of the image quality measurement algorithms based on Deep Learning. They also participated in the final prototype and validation tests for the code and configurations of the deep learning image quality module in \textsc{iquaflow}.

\bibliographystyle{unsrtnat}
\bibliography{template}  






\end{document}